%% file: paper925.tex
    \documentclass[runningheads]{llncs}
    \usepackage{wrapfig,lipsum,booktabs}
    \usepackage{graphicx}
    \usepackage{epsfig}
    \usepackage{svg}
    \usepackage{eucal}
    \usepackage{booktabs}
    \usepackage{todonotes}
    \usepackage{verbatim}
    \usepackage{amsmath}
    \usepackage{amssymb}
    \usepackage[normalem]{ulem}
    \usepackage{colortbl}
    \useunder{\uline}{\ul}{}

    \usepackage{multirow}
    \usepackage{caption}
    \usepackage{subcaption}
    \usepackage[colorlinks=true]{hyperref}
    \usepackage{orcidlink}
    
    \input{notation}
    
    \usepackage{todonotes}
    \usepackage{mathtools}

    \newcommand\und[1]{\underline{#1}}

    \newcommand{\xx}{\mathbf{x}}
    \newcommand{\yy}{\mathbf{y}}

    \newcolumntype{s}{>{\columncolor{gray!20}}c}

\newcommand\blfootnote[1]{%
  \begingroup
  \renewcommand\thefootnote{}\footnote{#1}%
  \addtocounter{footnote}{-1}%
  \endgroup
}

    \DeclareMathAlphabet{\mathcal}{OMS}{cmsy}{m}{n}
    
\begin{document}
    
    \title{Trust your neighbours:\\Penalty-based constraints for model calibration}

    \author{Balamurali Murugesan* \orcidlink{0000-0002-3002-5845}
    \and Sukesh Adiga Vasudeva \orcidlink{0000-0001-9754-1548}
    \and Bingyuan Liu
    \and Herve Lombaert \orcidlink{0000-0002-3352-7533}
    \and  Ismail Ben Ayed
    \and Jose Dolz \orcidlink{0000-0002-2436-7750}
    }
    
    \institute{ETS Montreal}
    
    \maketitle

    \newcommand{\mathbbm}[1]{\text{\usefont{U}{bbm}{m}{n}#1}}
    
    
    
    \begin{abstract}
    

    Ensuring reliable confidence scores from deep networks is of pivotal importance in critical decision-making systems, notably in the medical domain. While recent literature on calibrating deep segmentation networks has led to significant progress, 
    their uncertainty 
    is usually modeled by leveraging the information of individual pixels, which disregards the local structure of the object of interest. In particular, only the recent \textit{Spatially Varying Label Smoothing (SVLS)} approach addresses this issue by softening the pixel label assignments with a discrete spatial Gaussian kernel. In this work, we first present a constrained optimization perspective of SVLS and demonstrate that it enforces an implicit constraint on soft class proportions of surrounding pixels. Furthermore, our analysis shows that SVLS lacks a mechanism to balance the contribution of the constraint with the primary objective, potentially hindering the optimization process. 
    Based on these observations, 
    we propose a principled and simple solution based on equality constraints on the logit values, which enables to control explicitly both the enforced constraint and the weight of the penalty, offering more flexibility. Comprehensive experiments on a variety of well-known segmentation benchmarks demonstrate the superior 
    performance of the proposed approach. 
    The code is available at \url{https://github.com/Bala93/MarginLoss} \blfootnote{$^*$ Corresponding author}
    
    \end{abstract}
    
    
    \section{Introduction}
    
    Deep neural networks (DNNs) have achieved remarkable success in important areas of various domains, such as computer vision, machine learning and natural language processing. 
    Nevertheless, there exists growing evidence that suggests that these models are poorly calibrated, leading to overconfident predictions that may assign high confidence to incorrect predictions \cite{gal2016dropout,guo2017calibration}. 
    This represents a major problem, as inaccurate uncertainty estimates can have severe consequences in safety-critical applications such as medical diagnosis. 
    The underlying cause of network miscalibration is hypothesized to be the high capacity of these models, which makes them susceptible to overfitting on the negative log-likelihood 
    loss that is conventionally used during training \cite{guo2017calibration}.


    In light of the significance of this issue, there has been a surge in popularity for quantifying the predictive uncertainty in modern DNNs. A simple approach involves a post-processing step that modifies the softmax probability predictions of an already trained network \cite{ding2021local,guo2017calibration,Tomani2021Posthoc,zhang2020mix}. Despite its efficiency, this family of approaches presents important limitations, which include \textit{i)} a dataset-dependency on the value of the transformation parameters 
    and \textit{ii)} a large degradation observed under distributional drifts \cite{ovadia2019can}. A more principled solution integrates a term that penalizes confident output distributions into the learning objective, which explicitly maximizes the Shannon entropy of the model predictions during training \cite{pereyra2017regularizing}. Furthermore, findings from recent works on calibration \cite{mukhoti2020calibrating,muller2019does} have demonstrated that popular classification losses, such as Label Smoothing (LS) \cite{szegedy2016rethinking} and Focal Loss (FL) \cite{lin2017focal}, have a favorable effect
    on model calibration, as they implicitly integrate
    an entropy maximization objective. Following these works, \cite{liu2022devil,murugesan2022calibrating} presented a unified view 
    of state-of-the-art calibration approaches \cite{pereyra2017regularizing,szegedy2016rethinking,lin2017focal} 
    showing that these strategies can be viewed as approximations of a linear penalty imposing equality constraints on logit distances. The associated equality constraint results in gradients that continually push towards a non-informative solution, potentially hindering the ability to achieve the optimal balance between discriminative performance and model calibration. To alleviate this 
    limitation, \cite{liu2022devil,murugesan2022calibrating} proposed a simple and flexible alternative based on inequality constraints, which imposes a controllable margin on logit distances. Despite the progress brought by these methods, none of them explicitly considers pixel relationships, which is fundamental in the context of image segmentation.
    
    
    Indeed, the nature of structured predictions in segmentation, involves pixel-wise classification based on spatial dependencies, which limits the effectiveness of these strategies to yield performances similar to those observed in classification tasks. In particular, this potentially suboptimal performance can be attributed to the uniform (or near-to-uniform) distribution 
    enforced on the softmax/logits distributions, which disregards the spatial context information. To address this important issue, Spatially Varying Label Smoothing (SVLS) \cite{islam2021spatially} introduces a soft labeling approach that captures the structural uncertainty required in semantic segmentation. 
    In practice, smoothing the hard-label assignment is achieved through a Gaussian kernel applied across the one-hot encoded ground truth, which results in soft class probabilities based on neighboring pixels. Nevertheless, while the reasoning behind this smoothing strategy relies on the intuition of giving an equal contribution to the central label and all surrounding labels combined, its impact on the training, from an optimization standpoint, has not been studied. 
    
    The \textbf{contributions} of this work can be summarized as follows:

    \begin{itemize}
        \item We provide a constrained-optimization perspective of Spatially Varying Label Smoothing (SVLS) \cite{islam2021spatially}, demonstrating that it imposes an implicit constraint on a soft class proportion of surrounding pixels. Our formulation shows that SVLS lacks a 
        mechanism to control explicitly the importance of the constraint, which may hinder the optimization process as it becomes challenging to balance the constraint with the primary objective effectively.
    
        \item Following our observations, we propose a simple and flexible solution based on equality constraints on the logit distributions. The proposed constraint is enforced with a simple linear penalty, which incorporates an explicit mechanism to control the weight of the penalty. Our approach not only offers a more efficient strategy to model the logit distributions but implicitly decreases the logit values, which results in less overconfident predictions. 
    

        \item Comprehensive experiments 
        over multiple medical image segmentation benchmarks, including diverse targets and modalities, show the superiority of our method compared to state-of-the-art calibration losses.
        
    \end{itemize}
    

    \section{Methodology}


    \noindent \textbf{Formulation}. 
    Let us denote the training dataset as $\mathcal{D}(\mathcal{X}, \mathcal{Y})=\{(\xx^{(n)}, \yy^{(n)})\}_{n=1}^N$, with $\xx^{(n)} \in \mathcal{X} \subset \mathbb{R}^{\Omega_n}$ representing the $n^{th}$ image, $\Omega_n$ the spatial image domain, and $\yy^{(n)} \in \mathcal{Y} \subset \mathbb{R}^K$ its corresponding ground-truth label with $K$ classes, provided as a one-hot encoding vector. Given an input image $\xx^{(n)}$, a neural network parameterized by $\theta$ generates a softmax probability vector, defined as $f_{\theta}(\xx^{(n)})=\mathbf{s}^{(n)} \in \mathbb{R}^{\Omega_n \times K}$, where $\mathbf{s}$ is obtained after applying the softmax function over the logits $\rvl^{(n)} \in \mathbb{R}^{\Omega_n \times K}$. To simplify the notations, we omit sample indices, as this does not lead to any ambiguity. 
    

    \subsection{A constrained optimization perspective of SVLS}
    \label{ssec:SVLS}
    
    Spatially Varying Label Smoothing (SVLS) \cite{islam2021spatially} considers the surrounding class distribution of a given pixel $p$ in the ground truth $\yy$ to estimate the amount of smoothness over the one-hot label of that pixel. In particular, let us consider that we have a 2D patch $\bf{x}$ of size $d_1 \times d_2$ and its corresponding ground truth $\bf{y}$\footnote{For the sake of simplicity, we consider a patch as an image $\xx$ (or mask $\yy$), whose spatial domain $\Omega$ is equal to the patch size, i.e., $d_1 \times d_2$.}. Furthermore, the predicted softmax in a given pixel is denoted as $\rvs=[s_0,s_1,...,s_{k-1}]$. 
Let us now transform the surrounding patch of the segmentation mask around a given pixel 
into a unidimensional vector $\bf{y} \in \mathbb{R}^d$, where $d=d_1 \times d_2$. SVLS employs a discrete Gaussian kernel $\bf{w}$ to obtain soft class probabilities from one-hot labels, which can also be reshaped into $\bf{w} \in \mathbb{R}^d$. 
Following this, for a given pixel $p$, and a class $k$, SVLS \cite{islam2021spatially} can be defined as:
    
    
    \begin{align}
    \label{eq:svls}
    \tilde{y}^k_p = \frac{1}{| \sum_i^d w_i |} \sum_{i=1}^d y^k_i w_i .
    \end{align}
    
    Thus, once we replace the smoothed labels $\tilde{y}^k_p$ in the standard cross-entropy (CE) loss, the new learning objective becomes:
    
    \begin{align}
    \label{eq:svls-ce}
    \mathcal{L} 
    = - \sum_k \left(\frac{1}{|  \sum_i^d w_i |}  \sum_{i=1}^d y^k_i w_i \right) \log s^k_p ,
    \end{align}
    %
    %
    %
    where $s^k_p$ is the softmax probability for the class $k$ at pixel $p$ (the pixel in the center of the patch). Now, this loss can be decomposed into:
    
    \begin{align}
    \label{eq:svls-ce2}
    \mathcal{L}  =  - \frac{1}{| \sum_i^d w_i |} \sum_k   y^k_{p} \log s^k_p  - \frac{1}{|  \sum_i^d w_i |} \sum_k  
    \left(\sum_{\substack{i=1\\i\neq p}}^d y^k_i w_i\right) \log s^k_p ,
    \end{align}
    with $p$ denoting the index of the pixel in the center of the patch. 
    Note that the term in the left is the 
    cross-entropy between the posterior softmax probability and the hard label 
    assignment for pixel $p$. Furthermore, let us denote $\tau_k = \sum_{\substack{i=1\\i\neq p}}^d y^k_i w_i$ as the soft proportion of the class $k$ inside the patch/mask $\bf{y}$, weighted by the filter values $\mathbf{w}$. By replacing $\tau_k$ into the Eq. \ref{eq:svls-ce2}, and removing $|\sum_i^d w_i |$ as it multiplies both terms, 
    the loss becomes:
    
    


    \begin{align}
        \label{eq:tau}
        \mathcal{L}  = \underbrace{-\sum_k   y^k_{p} \log s^k_p}_{CE} \underbrace{- \sum_k \tau_k \log s^k_p}_{\textrm{Constraint on $\bm{\tau}$}}.
    \end{align}
    
    
    As $\bm{\tau}$ is constant, the second term in Eq. \ref{eq:tau} can be replaced by a Kullback-Leibler (KL) divergence, leading to the following learning objective:
    
    \begin{align}
        \label{eq:globalLoss}
        \mathcal{L}  \ceq \mathcal{L}_{CE} + 
        \mathcal{D}_{KL}(\bm{\tau}||\bf{s}) ,
    \end{align}
    where $\ceq$ stands for equality up to additive and/or non-negative multiplicative constant. Thus, optimizing the loss 
    in SVLS results in minimizing the 
    cross-entropy between the hard label and the softmax probability distribution on the pixel $p$, 
    while imposing the equality constraint $\bm{\tau}=\rvs$, where $\bm{\tau}$ depends on the class distribution of surrounding pixels. 
    Indeed, this term 
    implicitly enforces the softmax predictions to match the soft-class proportions computed around $p$. 

    \subsection{Proposed constrained calibration approach}
    \label{ssec:ours}
    
    Our previous analysis exposes two important limitations of SVLS: \textit{1)} the importance of the implicit constraint cannot be controlled explicitly, and \textit{2)} the prior $\bm{\tau}$ is derived from the $\sigma$ value in the Gaussian filter, making it difficult to model properly. 
    To alleviate this issue, we propose a simple solution, which consists in minimizing the standard cross-entropy between the softmax predictions and the one-hot encoded masks coupled with an explicit and controllable constraint on the logits $\rvl$. In particular, we propose to minimize the following constrained objective:

    \begin{align}
    \label{eq:proposed-constrained}
    \min \quad \mathcal{L}_{CE} \quad \textrm{s.t.} \quad  \bm{\tau} = \rvl, 
    \end{align}
    
    \noindent where $\bm{\tau}$ now represents a desirable prior, and $\bm{\tau} = \rvl $ is a hard constraint. Note that the reasoning behind working directly on the logit space is two-fold. First, observations in \cite{liu2022devil} suggest that directly imposing the constraints on the logits 
    results in better performance than in the softmax predictions. And second, by imposing a bounded constraint on the logits values\footnote{Note that the proportion priors are generally normalized.}, their magnitudes are further decreased, which has a favorable effect on model calibration \cite{muller2019does}. We stress that despite both \cite{liu2022devil} and our method enforce constraints on the predicted logits, \cite{liu2022devil} is fundamentally different. 
    In particular, \cite{liu2022devil} imposes an \textit{inequality} constraint on the logit distances so that it encourages uniform-alike distributions up to a given margin, disregarding the importance of each class in a given patch. 
    This can be important in the context of image segmentation, where 
    the uncertainty of a given pixel may be strongly correlated with the labels assigned to its neighbors. In contrast, our solution enforces \textit{equality} constraints on an adaptive prior, encouraging distributions close to class proportions in a given patch.

    Even though the constrained optimization problem presented in Eq. \ref{eq:proposed-constrained} could be solved by a standard Lagrangian-multiplier algorithm, we replace the hard constraint by a soft penalty of the form $\mathcal{P}(|\bm \tau - \rvl|)$, transforming our constrained problem into an unconstrained one, which is easier to solve. In particular, the soft penalty $\mathcal{P}$ should be a continuous and differentiable function that reaches its minimum when it verifies $\mathcal{P}(|\bm \tau - \rvl|) \geq \mathcal{P}(\bm 0),  \, \forall \, \bm l \in \mathbb{R}^{K}$, i.e., when the constraint is satisfied. Following this, when the constraint $|\bm \tau - \rvl|$ deviates from $\bm 0$ the value of the penalty term increases. Thus, we can approximate the problem in Eq. \ref{eq:proposed-constrained} as the following simpler unconstrained problem:
    

    \begin{align}
    \label{eq:proposed-unconstrained}
    \min \quad \mathcal{L}_{CE} + \lambda \sum_k |\tau_k - l_k| ,
    \end{align}
    where the penalty is modeled here as a ReLU function, whose importance is controlled by the hyperparameter $\lambda$.
    
    
    
    
    
    
    \section{Experiments}
    
    \subsection{Setup}

    
    \noindent \textbf{Datasets.} \textbf{
    FLARE Challenge} \cite{AbdomenCT-1K} contains $360$ volumes of multi-organ abdomen CT with their corresponding pixel-wise masks, which are resampled to a common space and cropped to 192$\times$192$\times$30. 
    \textbf{
    ACDC Challenge} \cite{bernard2018deep} consists of 100 patient exams containing cardiac MR volumes and their respective 
    segmentation masks. 
    Following the standard practices on this dataset, 2D slices are extracted from the 
    volumes and resized to 224$\times$224. \textbf{
    BraTS-19 Challenge}~\cite{Menze2015TheBRATSJ,Bakas2017AdvancingFeaturesJ,Bakas2018IdentifyingChallengeJ} contains $335$ multi-modal MR scans (FLAIR, T1, T1-contrast, and T2) with their corresponding segmentation masks, where each volume of dimension 155$\times$240$\times$240 is resampled to 128$\times$192$\times$192. More details about these datasets, such as the train, validation and testing splits, can be found in Supp. Material. 

     \noindent \textbf{Evaluation metrics.} To assess the discriminative performance of the evaluated models, we resort to standard segmentation metrics in medical segmentation, which includes the DICE coefficient (DSC) and the 95\% Hausdorff Distance (HD). 
    To evaluate the calibration performance, we employ the expected calibration error (ECE) \cite{naeini2015obtaining} on foreground classes, as in \cite{islam2021spatially}, and classwise expected calibration error (CECE) \cite{kull2019beyond}, following \cite{mukhoti2020calibrating,murugesan2022calibrating} (more details in Supp. Material). 
    
    \noindent \textbf{Implementation details}. We benchmark the proposed model against several losses, including state-of-the-art calibration losses. 
    These models include 
    the compounded CE + Dice loss (CE+DSC), 
    FL \cite{lin2017focal}, Entropy penalty (ECP) \cite{pereyra2017regularizing}, 
    LS \cite{szegedy2016rethinking}, SVLS \cite{islam2021spatially} and MbLS \cite{liu2022devil}. Following the literature, we consider the hyperparameters values typically employed and select the value which provided the best average DSC on the validation set across all the datasets. More concretely, for FL, $\gamma$ values of 1, 2, and 3 are considered, whereas 0.1, 0.2, and 0.3 are used for $\alpha$ and $\lambda$ in LS and ECP, respectively. We consider the margins of MbLS to be 3, 5, and 10, while fixing $\lambda$ to 0.1, as in \cite{murugesan2022calibrating}. In the case of SVLS, the one-hot label smoothing is performed with a kernel size of 3 and $\sigma=[0.5,1,2]$. For training, we fixed the batch size to 16, epochs to 100, and used ADAM \cite{kingma2014adam}, with a learning rate of 10$^{-3}$ for the first 50 epochs, and reduced to 10$^{-4}$ afterwards. Following \cite{murugesan2022calibrating}, the models are trained on 2D slices, and the evaluation is performed over 3D volumes. Last, we use the following 
    prior 
    $\tau_k = \sum_{\substack{i=1}}^d y^k_i$, which is computed over a 3$\times$3 patch, similarly to SVLS.

    \subsection{Results}
    
    \noindent \textbf{Comparison to state-of-the-art.} Table \ref{table:main} reports the discriminative and calibration results achieved by the different methods. We can observe that, across all the datasets, the proposed method consistently outperforms existing approaches, always ranking first and second in all the metrics. Furthermore, while other methods may obtain better performance than the proposed approach in a single metric, their superiority strongly depends on the selected dataset. For example, ECP \cite{pereyra2017regularizing} yields very competitive performance on the FLARE dataset, whereas it ranks among the worst models in ACDC or BraTS. 
    
    
    
    \begin{table}[t]
    
    \centering
    \scriptsize
    \caption{\textbf{Comparison to state-of-the-art.} Discriminative (DSC $\uparrow$, HD $\downarrow$) and calibration (ECE $\downarrow$, CECE $\downarrow$) performance obtained by the different models (best method in bold, and second best in bold and underlined).}
    \begin{tabular}{l | cccc | cccc | cccc}
                        \toprule
                        & \multicolumn{4}{c|}{FLARE} & 
                        \multicolumn{4}{c|}{ACDC} & \multicolumn{4}{c}{BraTS}  \\
                        \midrule
                        & DSC & HD & ECE & CECE & DSC & HD & ECE & CECE & DSC & HD & ECE & CECE                      \\
        \midrule
    CE+DSC ($\lambda=1$) & 0.846 & 5.54 & 0.058 & 0.034 &  \textbf{\underline{0.828}} & 3.14 & 0.137 & 0.084 & 0.777 & \textbf{\underline{6.96}} & 0.178 & 0.122 \\
    FL \cite{lin2017focal} ($\gamma=3$)  &  0.834 & 6.65 & 0.053 & 0.059 & 0.620 & 7.30 & 0.153 & 0.179 & \textbf{\underline{0.848}} & 9.00 & \textbf{0.097} & 0.119 \\
    ECP \cite{pereyra2017regularizing} ($\lambda=0.1$)   & \textbf{\underline{0.860}} & \textbf{\underline{5.30}} & \textbf{\underline{0.037}} &  \textbf{0.027} & 0.782 & 4.44 & 0.130  & 0.094 & 0.808 & 8.71 & 0.138 & 0.099 \\
    LS \cite{szegedy2016rethinking} ($\alpha=0.1$)   & \textbf{\underline{0.860}} & 5.33 & 0.055 & 0.049 &  0.809 & 3.30 & \textbf{\underline{0.083}}  & 0.093  & 0.820 & 7.78 & \textbf{\underline{0.112}} & 0.108 \\
    SVLS \cite{islam2021spatially} ($\sigma=2$)  & 0.857 & 5.72 & 0.039 & 0.036 & 0.824 & \textbf{\underline{2.81}} & 0.091  & 0.083  & 0.801 & 8.44 & 0.146 & 0.111 \\
    MbLS \cite{liu2022devil} ($m$=5)
    & 0.836 & 5.75 & 0.046 & 0.041  & 0.827 & 2.99 & 0.103 & \textbf{\underline{0.081}} & 0.838 & 7.94 & 0.127 & \textbf{0.095} \\
    Ours ($\lambda=0.1)$ & \textbf{0.868} & \textbf{4.88}  & \textbf{0.033} &  \textbf{\underline{0.031}}  & \textbf{0.854} & \textbf{2.55} & \textbf{0.048} & \textbf{0.061} & \textbf{0.850} & \textbf{5.78} & \textbf{\underline{0.112}} & \textbf{\underline{0.097}} \\
    \bottomrule
    \end{tabular}
   
    \label{table:main}
    \end{table}

\begin{figure}[t!]

         \centering
         \begin{subfigure}[b]{0.485\linewidth}
             \centering
             \includegraphics[width=\linewidth]{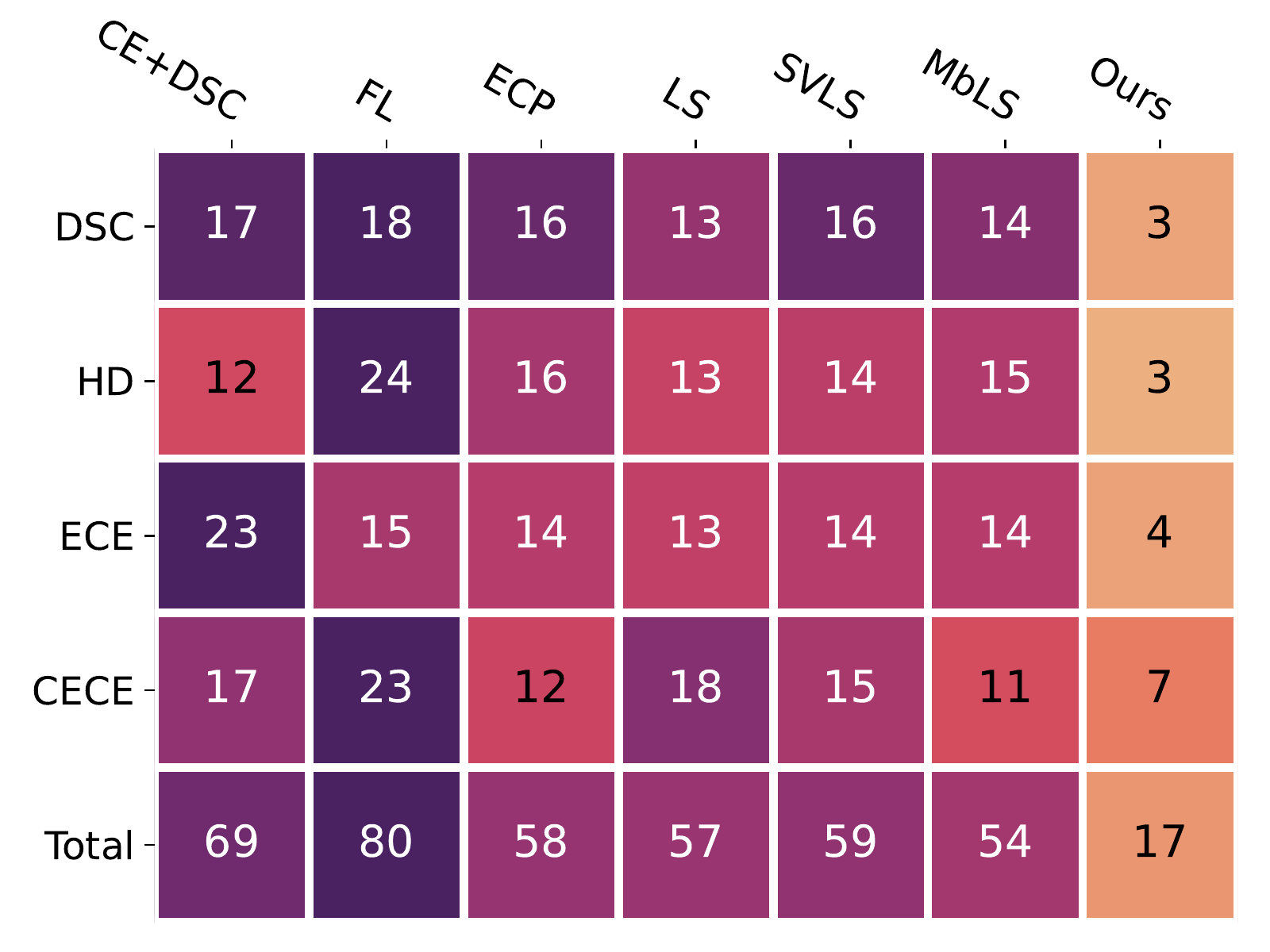}
         \end{subfigure}
         \begin{subfigure}[b]{0.485\linewidth}
             \centering
             \includegraphics[width=\linewidth]{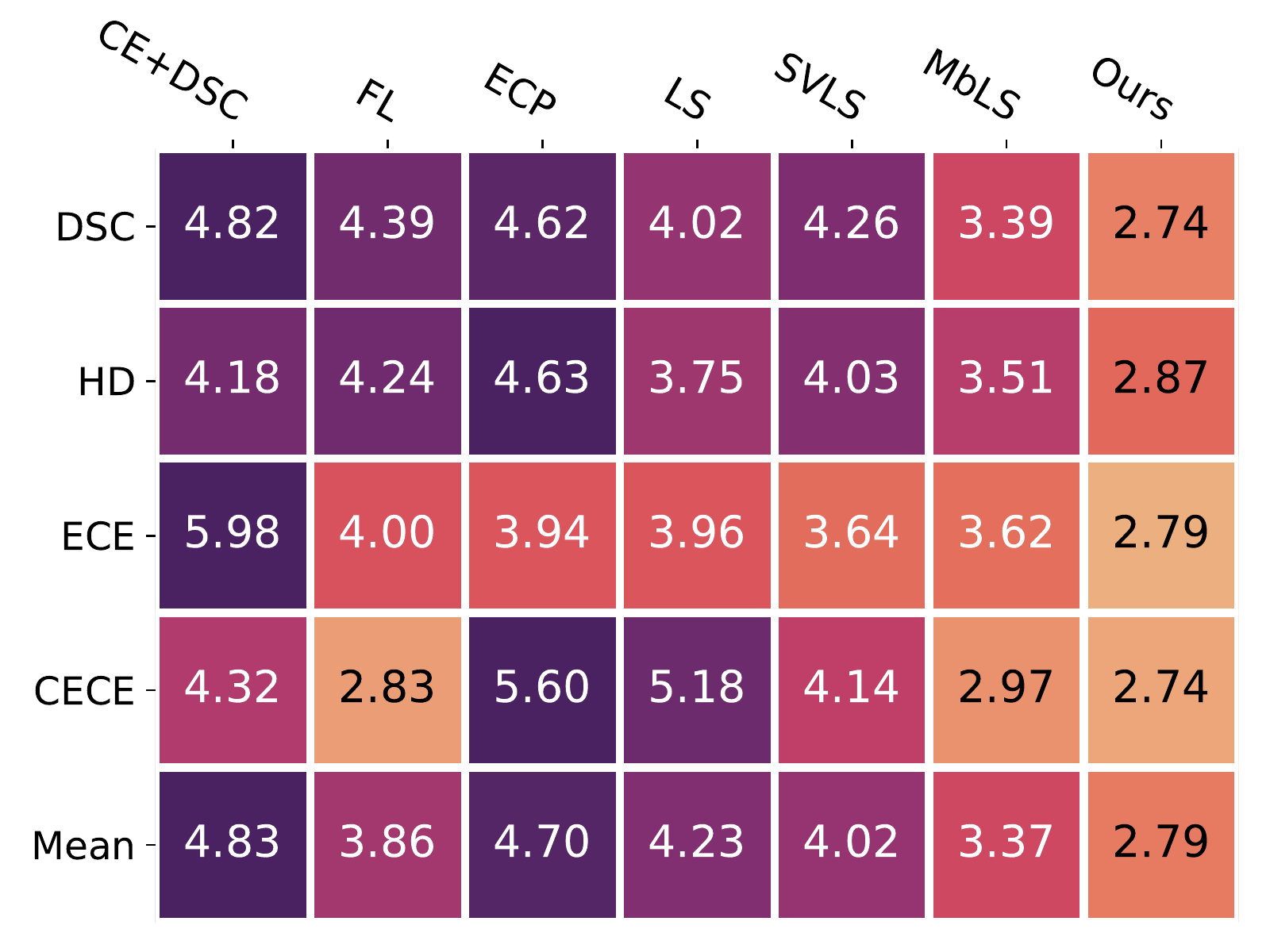}
         \end{subfigure}
        \caption{\textbf{Sum-rank and mean-rank evaluation}. Ranking of the different methods based on the sum-rank (\textit{left}) and mean of case-specific (\textit{right}) approaches. The lower the value, the better the performance.}
        \label{fig:main-sumRank}
    \end{figure}

    To have a better overview of the performance of the different methods, 
    we follow the evaluation strategies adopted in several MICCAI Challenges, i.e., sum-rank \cite{mendrik2015mrbrains} and mean-case-rank \cite{MAIER2017250}. As we can observe in the heatmaps provided in Fig. \ref{fig:main-sumRank}, our approach yields the best rank across all the metrics in both strategies, clearly outperforming any other method. Interestingly, some methods such as FL or ECP typically provide well-calibrated predictions, but at the cost of degrading their discriminative performance. 

   

   

    \noindent \textbf{Ablation studies.} \textbf{1-Constraint over logits \textit{vs} softmax.} Recent evidence \cite{liu2022devil} suggests that imposing constraints on the logits presents a better alternative than its softmax counterpart. To demonstrate that this observation holds in our model, we present the results of our formulation 
    when the constraint is enforced on the softmax distributions, i.e., replacing $\bf{l}$ by $\bf{s}$ (Table \ref{table:ablation}, \textit{top}), which yields inferior results. \textbf{2-Choice of the penalty.} To solve the unconstrained problem in Eq. \ref{eq:proposed-unconstrained}, we can approximate the second term with a liner penalty, modeled as a ReLU function. Nevertheless, we can resort to other polynomial penalties, e.g., quadratic penalties, whose main difference stems from the more aggressive behavior of quadratic penalties over larger constraint violations. The results obtained when the linear penalty is replaced by a quadratic penalty are reported in Table \ref{table:ablation} (\textit{middle}). 
    From these results, we can observe that, while a quadratic penalty could achieve better results in a particular dataset (e.g., ACDC or calibration performance on BraTS), a linear penalty yields more consistent results across datasets. \textbf{3-Patch size.} For a fair comparison with SVLS, we used a patch of size 3~$\times$~3 in our model. Nevertheless, we now investigate the impact of employing a larger patch to define the prior $\bm \tau$, whose results are presented in Table \ref{table:ablation} (\textit{bottom}). Even though a larger patch seems to bring comparable results in one dataset, the performance on the other two datasets is largely degraded, which potentially hinders its scalability to other applications. We believe that this is due to the higher degree of noise in the class distribution, particularly when multiple organs overlap, as the employed patch covers a wider region.

 \begin{table}[h!]
    \centering
    \scriptsize
    \caption{Empirical results to motivate our methodological and technical choices.}
    \begin{tabular}{l | cccc | cccc | cccc}
                        \toprule
                        & \multicolumn{4}{c|}{FLARE} & 
                        \multicolumn{4}{c|}{ACDC} & \multicolumn{4}{c}{BraTS}  \\
                        \midrule
                        & DSC & HD & ECE & CECE & DSC & HD & ECE & CECE & DSC & HD & ECE & CECE \\
        \midrule
    Constraint on $\bf{s}$ & 0.862& 5.14& 0.043& 0.030 & 0.840  & 2.66  & 0.068  & 0.071 & 0.802& 8.28& 0.145& 0.104 \\
    L2-penalty & 0.851 & 5.48 & 0.065 & 0.054 & 0.871 & 1.78 & 0.059 & 0.080 & 0.851 & 7.90 & 0.078 & 0.091 \\
    Patch size: 5 $\times$ 5 & 0.875 & 5.96 & 0.032 & 0.031  & 0.813 & 3.50 & 0.078 & 0.077  & 0.735  & 7.45 & 0.119 & 0.092 \\
    \bottomrule
    \end{tabular}
    \label{table:ablation}
    \end{table}

    
    \noindent \textbf{Impact of the prior.} A benefit of the proposed formulation 
    is that diverse 
    priors can be enforced on the logit distributions. Thus, we now assess the impact of different priors $\bm \tau$ in our formulation 
    (See Supplemental Material for a detailed explanation). 
    The results presented in Table \ref{table:prior} reveal that selecting a suitable prior can further improve the performance of our model.

    \begin{table}[t]
    \centering
    \scriptsize
    \caption{Impact of using different priors ($\bm \tau$) in Eq. \ref{eq:proposed-unconstrained}.}
    \begin{tabular}{l | cccc | cccc | cccc}
                        \toprule
                        & \multicolumn{4}{c|}{FLARE} & 
                        \multicolumn{4}{c|}{ACDC} & \multicolumn{4}{c}{BraTS}  \\
                        \midrule
                        Prior $\bm \tau $& DSC & HD & ECE & CECE & DSC & HD & ECE & CECE & DSC & HD & ECE & CECE \\
        \midrule
    Mean & \textbf{0.868} & \textbf{4.88}  & \textbf{0.033} &  \textbf{0.031} & 0.854 & 2.55 & 0.048 & 0.061 & \textbf{0.850} & \textbf{5.78} & 0.112 & 0.097 \\
    Gaussian & 0.860 & 5.40 & \textbf{0.033} & \textbf{\und{0.032}} & \textbf{\und{0.876}} & 2.92 & \textbf{\und{0.042}} & \textbf{0.053} & 0.813 & \textbf{\und{7.01}} & 0.140 & 0.106 \\
     Max & 0.859 & \textbf{\und{4.95}} & 0.038 & 0.036 & \textbf{\und{0.876}} & \textbf{\und{1.74}} & 0.046 & \textbf{\und{0.054}} & 0.833 & 8.25 & 0.114 & 0.094 \\
     Min & 0.854 & 5.42 & \textbf{\und{0.034}} & 0.033 & \textbf{0.881} & 1.80 & \textbf{0.040} & \textbf{0.053} & 0.836 & 7.23 & \textbf{\und{0.104}} & \textbf{\und{0.092}} \\
     Median & \textbf{\und{0.867}} & 5.90 & \textbf{0.033} & \textbf{\und{0.032}} & 0.835 & 3.29 & 0.075 & 0.075 & \textbf{\und{0.837}} & 7.53 & \textbf{0.095} & \textbf{0.089} \\
     Mode & 0.854 & 5.41 & 0.035 & 0.034 & \textbf{\und{0.876}} & \textbf{1.62} & 0.045 & 0.056 &  0.808 & 8.21 & 0.135 & 0.113 \\
    \bottomrule
    \end{tabular}
    \label{table:prior}
    \end{table}

\begin{figure}[h!]
         \centering
         \begin{subfigure}[b]{0.328\linewidth}
             \centering
             \includegraphics[width=\linewidth,trim={0.2cm 0.2cm 0.4cm 0.1cm},clip]{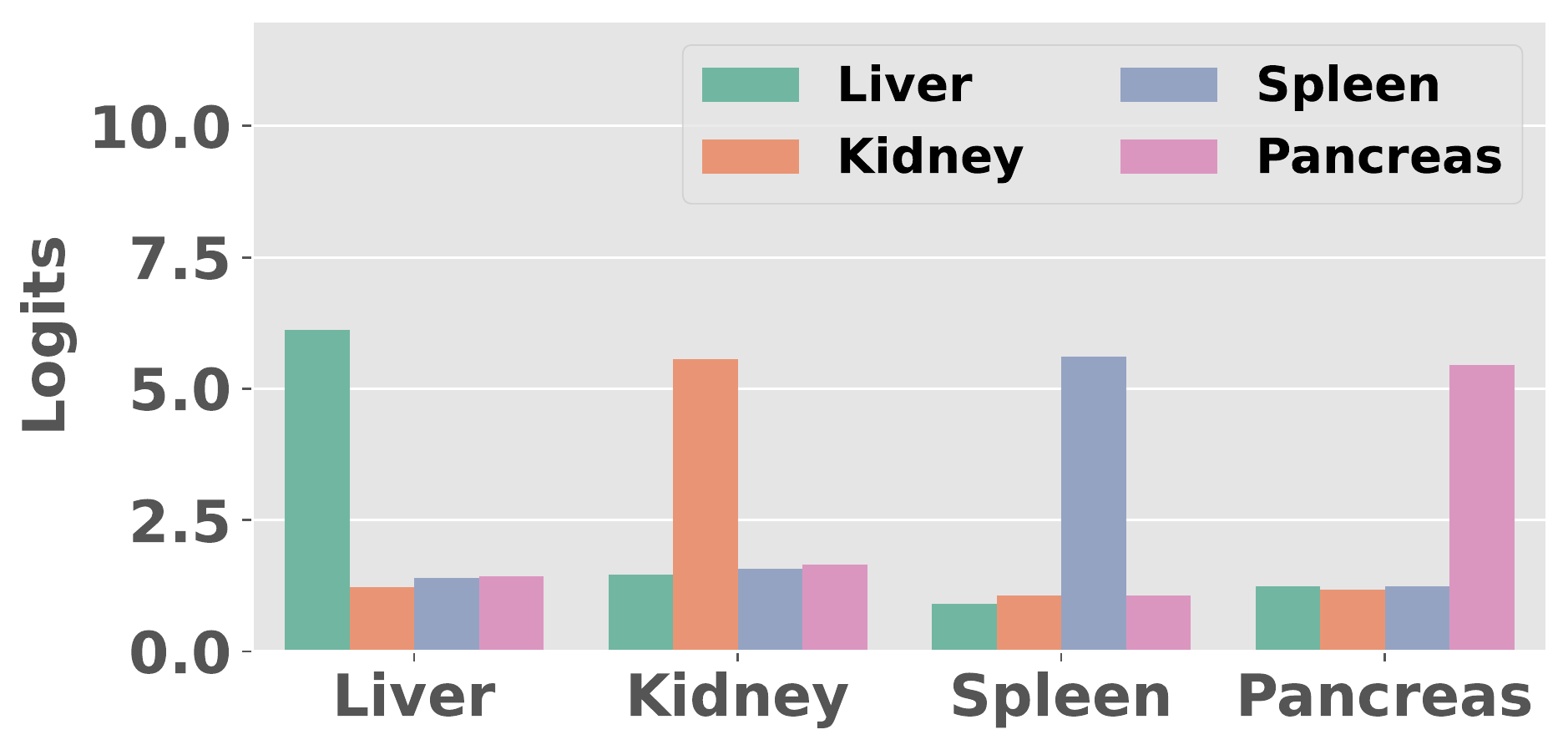}
         \end{subfigure}
         \begin{subfigure}[b]{0.328\linewidth}
             \centering
             \includegraphics[width=\linewidth,trim={0.2cm 0.2cm 0.4cm 0.1cm},clip]{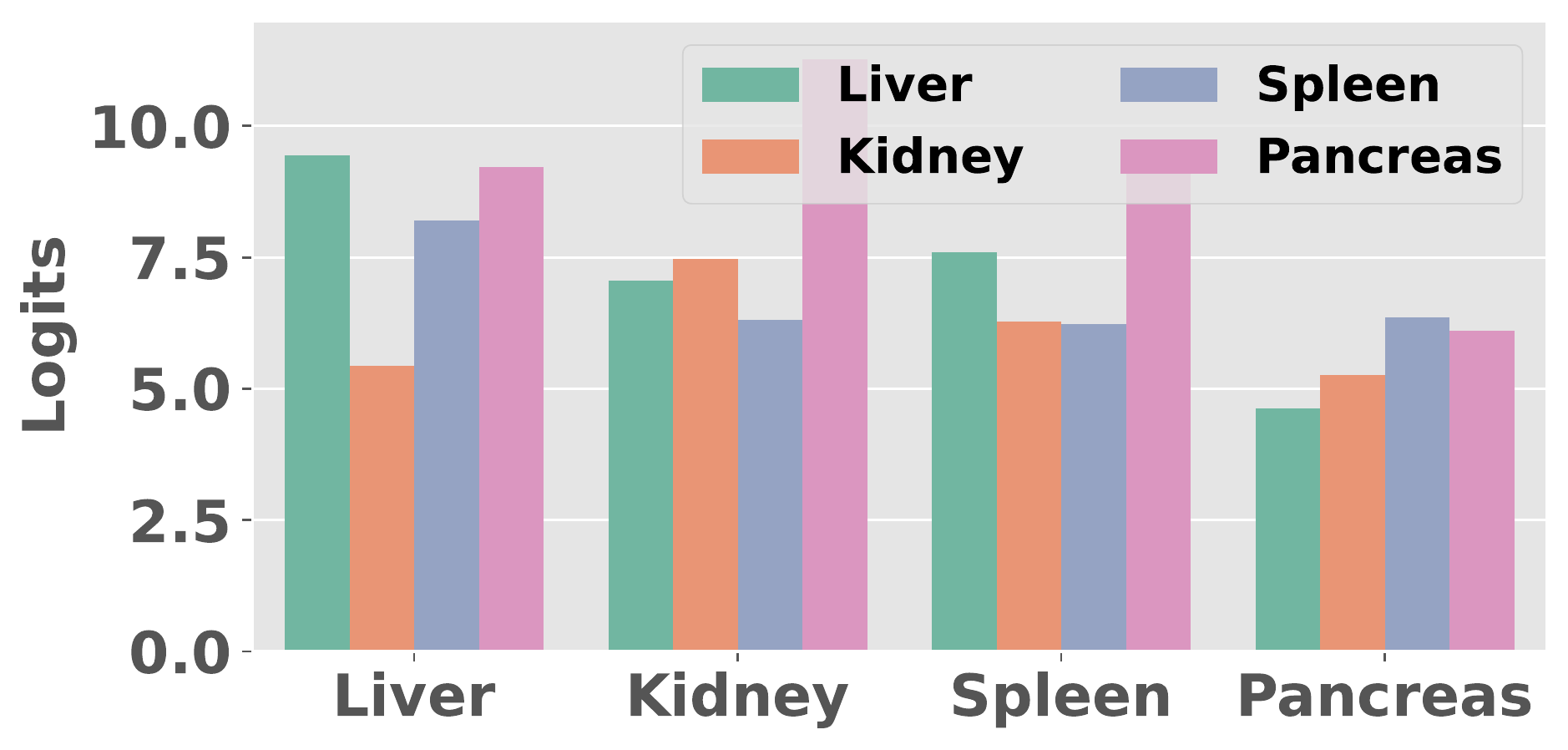}
         \end{subfigure}
         \begin{subfigure}[b]{0.328\linewidth}
             \centering
             \includegraphics[width=\linewidth,trim={0.2cm 0.2cm 0.4cm 0.1cm},clip]{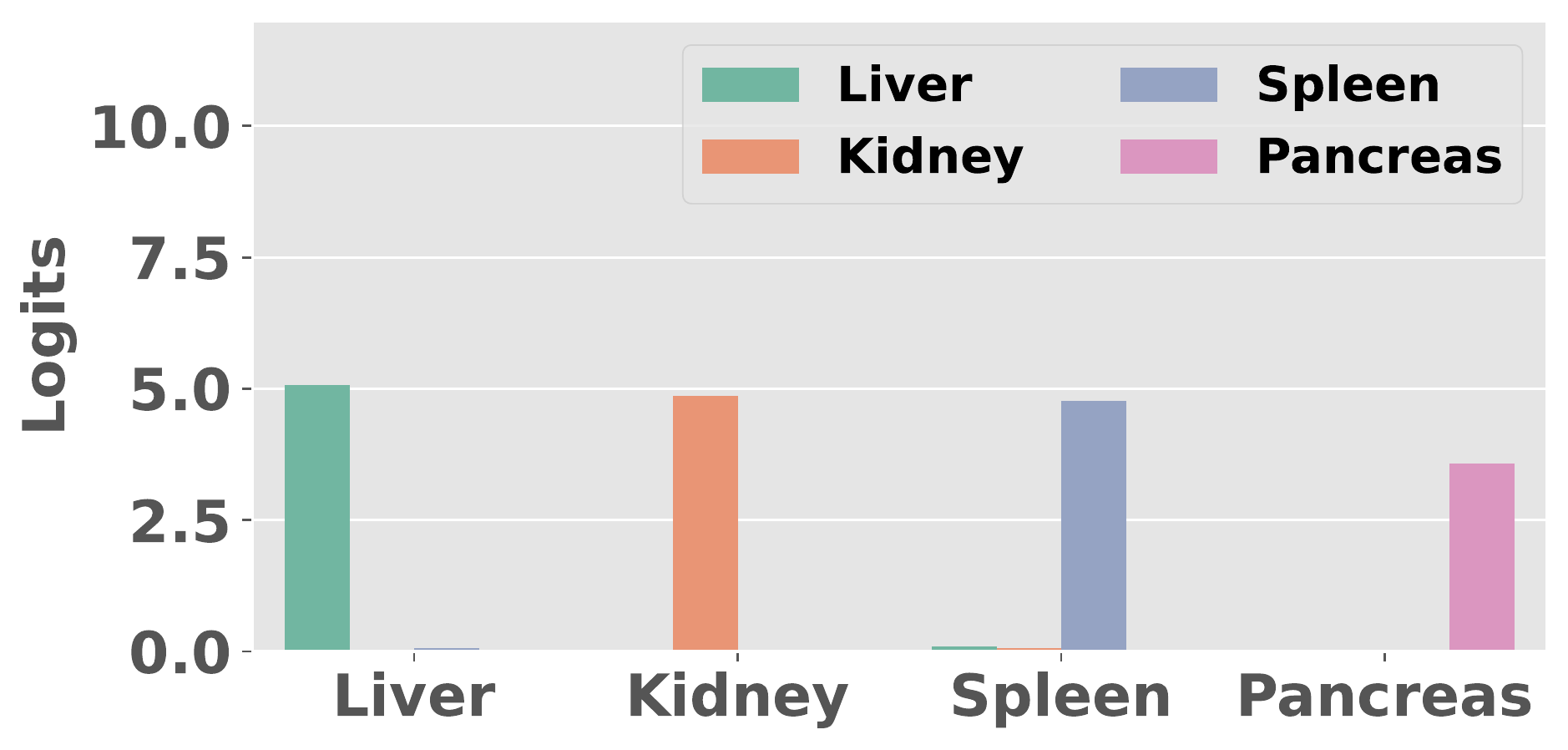}
         \end{subfigure}
        \caption{\textbf{Distribution of logit values.} From left to right: MbLS, SVLS and ours.}
        \label{fig:logit-dist}
    \end{figure}

    \noindent \textbf{Magnitude of the logits.} 
    To empirically demonstrate that the proposed solution 
    decreases the logit values, we plot average logit distributions across classes on the FLARE test set (Fig. \ref{fig:logit-dist}). In particular, we first separate all the voxels based on their ground truth labels. Then, for each category, we average the per-voxel vector of logit predictions (in absolute value). We can observe that, compared to SVLS and MbLS, --which also imposes constraints on the logits--, our approach leads to much lower logit values, particularly compared to SVLS. 
    
    \section{Conclusion}
    We have presented a constrained-optimization perspective of SVLS, which has revealed two important limitations of this method. First, the implicit constraint enforced by SVLS cannot be controlled explicitly. And second, the prior imposed in the constraint is directly derived from the Gaussian kernel used, which makes it hard to model. In light of these observations, we have proposed a simple alternative based on equality constraints on the logits, which allows to control the importance of the penalty explicitly, and the inclusion of any desirable prior in the constraint. Our results suggest that the proposed method 
    improves the quality of the uncertainty estimates, while enhancing the segmentation performance. 
    
     
    \bibliographystyle{splncs04}
    \bibliography{paper925}


    \newpage
    \setcounter{page}{1}
    \setcounter{table}{0}
    \setcounter{section}{0}
    \setcounter{equation}{0}
      
 
    
\begin{table}[h!]
\centering
\scriptsize
\caption{Additional details on the different datasets employed in our evaluation.}
\begin{tabular}{l|ccc|l}
                    \toprule
                    & \multicolumn{3}{c|}{Dataset splits} &  \multicolumn{1}{c}{Classes}
                      \\
                    \midrule
                    & Train & Val & Test & \\
    \midrule
FLARE & 240& 40& 80& (1) Liver, (2) Kidneys, (3) Spleen, (4) Pancreas \\
ACDC & 70& 10& 20& (1) Left Ventricle (LV), (2) Right Ventricle (RV), (3) Myocardium (MYO) \\
BraTS & 235 & 35 & 65 & (1) Tumor Core (TC), (2) Enhancing Tumor (ET), (3) Whole Tumor (WT) \\
\bottomrule
\end{tabular}
\label{table:dataset}
\end{table}

We provide the formulation of different metrics used for evaluation:

\noindent \textbf{- Expectation Calibration Error (ECE).} The ECE can be approximated as a weighted average of the absolute difference between the accuracy and confidence of each bin: $ECE = \sum_{i=1}^{M}\frac{|B_{i}|}{N}|A_{i} - C_{i}|$, where $M$ denotes the number of equispaced bins, $B_{i}$ denote the set of samples with confidences belonging to the $i^{th}$ bin, $A_{i}$ is the accuracy of the $i$-th bin, and it is computed as $A_{i}=\frac{1}{|B_{i}|}\sum_{j \in B_{i}}1(\hat{y_{j}} =y_{j})$, where 1 is the indicator function, $\hat{y_{j}}$, and $y_{j}$ are the predicted and ground-truth labels for the $j^{th}$ sample. Similarly, the confidence $C_{i}$ of the $i^{th}$ bin is computed as $C_{i}=\frac{1}{|B_{i}|}\sum_{j \in B_{i}}\hat{p}_{j}$, i.e. $C_{i}$ is the average confidence of all samples in the bin.

\noindent \textbf{- Classwise ECE.} The simple classwise extension of the ECE metric is defined as: $CECE = \sum_{i=1}^{M}\sum_{j=1}^{K}\frac{|B_{i,j}|}{N}|A_{i,j} - C_{i,j}|$, where $K$ is the number of classes, $B_{ij}$ denotes the set of samples from the $j^{th}$ class in the $i^{th}$ bin, $A_{i,j}=\frac{1}{|B_{i,j}|}\sum_{k \in B_{i,j}}1(j =y_{k})$ and $C_{i,j}=\frac{1}{|B_{ij}|}\sum_{k \in B_{i,j}}\hat{p}_{kj}$.

\noindent \textbf{- Sum-rank.} We follow the strategy followed in several MICCAI Challenges, e.g., MRBrainS \cite{mendrik2015mrbrains}, where the final ranking is given as the sum of individual ranking metrics: $r_{mc}$, where $r_{mc}$ ${ R_{T} = \sum_{m=0}^{|M|}  r_{m}}$, where $r_{m}$ is the rank of the segmentation model for the metric $m$ (mean).

\noindent \textbf{-Mean case-rank.} Furthermore, to account for the different complexities of each sample, we follow the mean-case-rank strategy, which has been employed in other MICCAI Challenges, e.g., \cite{MAIER2017250}. We first compute the DSC, HD, ECE, and CECE values for each sample, and establish each method’s rank based on these metrics, separately for each case. Then, we compute the mean rank over all four evaluation metrics, per case, to obtain the method’s rank for that given sample. Finally, we compute the mean over all case-specific ranks to obtain the method’s final rank.
    
Let the $\yy^{*} \in \mathbb{R}$ be the label encoding with the unique values $\{1, ..K\}$ for one-hot labels $\yy \in \mathbb{R}^{K}$. For each of the prior, the label $\yy$ is updated to $\tilde\yy \in \mathbb{R}^{K}$. Thus, for a pixel ${p}$ with patch $d = d_{1} \times d_{2}$, the following equations can be used to obtain the prior:

\begin{equation}
    \textrm{Mean:} \quad \tilde{y}^{k}_{p} = \frac{1}{d} \sum_{i=1}^d y_{i}^{k}
\end{equation}

\begin{equation}
    \textrm{Gaussian:} \quad  \tilde{y}^k_p = \frac{1}{| \sum_i^d w_i |} \sum_{i=1}^d y^k_i w_i,
\end{equation}
where $w_{i}$ is the Gaussian kernel.

\begin{equation}
    \textrm{Max:} \quad \tilde{y}^{k}_{p} = \begin{cases} 1,& \underset{d}{max}(y^{*k}) == k \\ 0,& \textrm{otherwise}. \end{cases}
\end{equation}

The priors pertaining to Min, Mode, Median can also be obtained by replacing Eq. 3 with respective to order statistics operation. 






\begin{table*}[h!]
\tiny
\centering
\label{tab:main-disc}
\caption{Class-wise segmentation scores on FLARE, ACDC, and BraTS datasets.}
\begin{tabular}{s|c|cccccccccccccc}
\toprule
&  Region & \multicolumn{2}{c}{CE+DSC} & \multicolumn{2}{c}{FL} & \multicolumn{2}{c}{ECP} & \multicolumn{2}{c}{LS} & \multicolumn{2}{c}{SVLS} & \multicolumn{2}{c}{MbLS} & \multicolumn{2}{c}{Ours}\\
\midrule
& & DSC & HD & DSC & HD & DSC & HD & DSC & HD & DSC & HD & DSC & HD & DSC & HD \\

 \midrule
 & Liver & 0.942 & 7.60 & 0.942 & 7.54 & 0.953 & 7.41 & 0.952 & 8.50 & 0.951 & 7.72 & 0.941 & 7.18 & 0.954 & 6.04\\
 & Kidney & 0.941 & 2.43 & 0.942 & 2.16 & 0.950 & 2.05 & 0.947 & 1.76 & 0.947 & 1.84 & 0.937 & 2.49 & 0.952 & 1.84\\
 & Spleen & 0.867 & 3.70 & 0.875 & 9.09 & 0.887 & 3.98 & 0.905 & 4.62 & 0.879 & 6.40 & 0.868 & 4.73 & 0.900 & 4.26\\
 & Pancreas & 0.634 & 8.42 & 0.578 & 7.80 & 0.649 & 7.77 & 0.637 & 6.45 & 0.650 & 6.91 & 0.596 & 8.61 & 0.664 & 7.37\\
 \rowcolor{gray!20} \multirow{-5}{*}{\rotatebox[origin=c]{90}{FLARE}} & Mean & 0.846 & 5.54 & 0.834 & 6.65 & 0.860 & 5.30 & 0.860 & 5.33 & 0.857 & 5.72 & 0.836 & 5.75 & 0.868 & 4.88\\

 \midrule
 & RV 
 & 0.799 & 3.10 & 0.580 & 9.37 & 0.751 & 4.93 & 0.796 & 3.34 & 0.791 & 2.89 & 0.812 & 2.59 & 0.837 & 3.02 \\ 
 & MYO & 0.795 & 2.57 & 0.557 & 5.55 & 0.757 & 3.54 & 0.772 & 3.07 & 0.798 & 2.66 & 0.795 & 2.86 & 0.820 & 2.04 \\ 
 & LV & 0.889 & 3.75 & 0.724 & 6.97 & 0.839 & 4.85 & 0.858 & 3.49 & 0.882 & 2.89 & 0.875 & 3.53 & 0.905 & 2.59 \\ 
 \rowcolor{gray!20} \multirow{-4}{*}{\rotatebox[origin=c]{90}{ACDC}} & Mean & 0.828 & 3.14 & 0.620 & 7.30 & 0.782 & 4.44 & 0.809 & 3.30 & 0.824 & 2.81 & 0.827 & 2.99 & 0.854 & 2.55\\
 
 \midrule
 & TC & 0.730 & 5.73 & 0.799 & 7.80 & 0.749 & 7.53 & 0.773 & 5.16 & 0.744 & 7.56 & 0.803 & 4.88 & 0.804 & 3.98\\ 
 & ET & 0.746 & 8.27 & 0.854 & 10.02 & 0.790 & 11.31 & 0.807 & 10.23 & 0.783 & 9.22 & 0.821 & 10.85 & 0.854 & 6.58\\
 & WT & 0.855 & 6.88 & 0.889 & 9.19 & 0.884 & 7.28 & 0.879 & 7.94 & 0.877 & 8.55 & 0.889 & 8.09 & 0.893 & 6.78\\ 
 \rowcolor{gray!20} \multirow{-4}{*}{\rotatebox[origin=c]{90}{BraTS}} & Mean & 0.777 & 6.96 & 0.848 & 9.00 & 0.808 & 8.71 & 0.820 & 7.78 & 0.801 & 8.44 & 0.838 & 7.94 & 0.850 & 5.78\\  
\bottomrule
\end{tabular}
\end{table*}


\begin{figure}
    \centering
    \includegraphics[width=\linewidth]{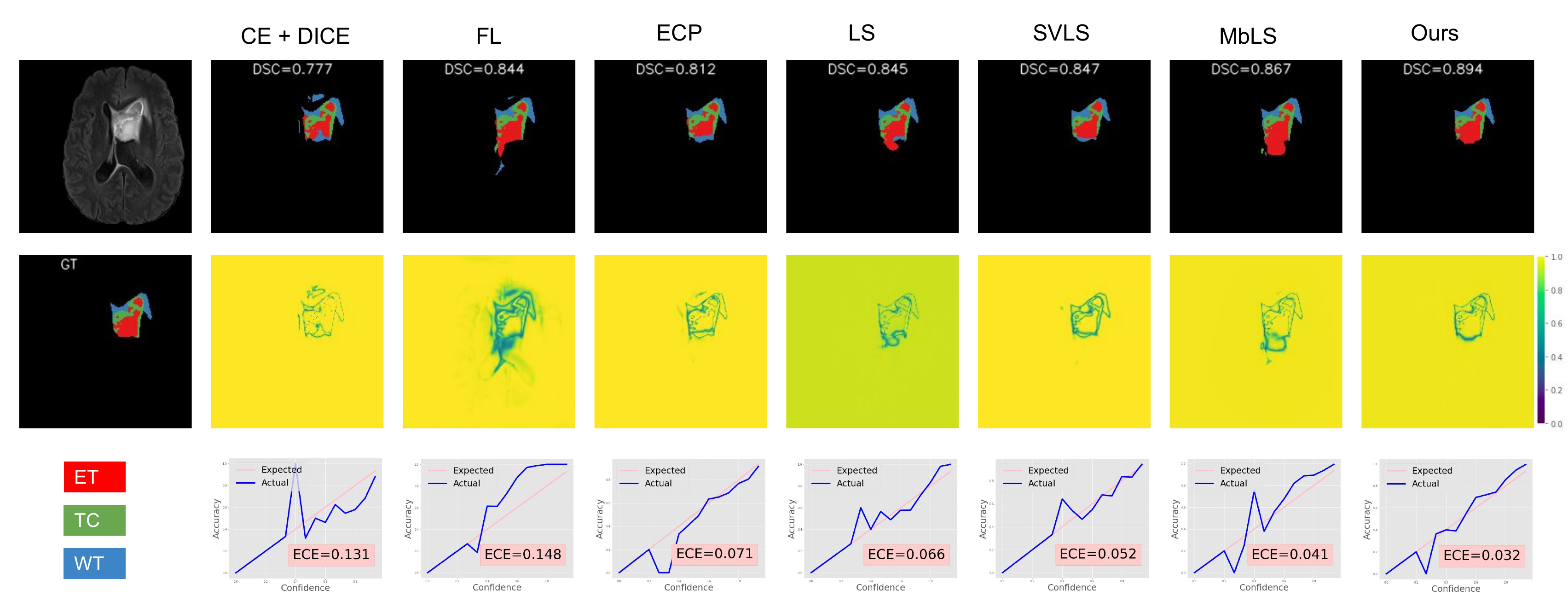}
    \caption{Qualitative results on BraTS dataset for different methods. In particular, we show the original image and the corresponding segmentation masks provided by each method (\textit{top row}), the ground-truth (GT) mask followed by maximum confidence score of each method (\textit{middle row}) and the respective reliability plots (\textit{bottom row}). Methods from left to right: CE+DICE, FL, ECP, LS, SVLS, MbLS, Ours}.
\end{figure}
    
\end{document}

%% file: notation.tex
\usepackage{amsmath,amsfonts,bm}

\newcommand{\ceq}{\stackrel{\mathclap{\normalfont\mbox{c}}}{=}}

\def\eqref#1{equation~\ref{#1}}

\def\1{\bm{1}}

\def\rvl{{\mathbf{l}}}

\def\rvs{{\mathbf{s}}}

\DeclareMathAlphabet{\mathsfit}{\encodingdefault}{\sfdefault}{m}{sl}
\SetMathAlphabet{\mathsfit}{bold}{\encodingdefault}{\sfdefault}{bx}{n}

